\documentclass{ifacconf}
\usepackage{graphicx}      
\usepackage{natbib}        
\usepackage{tikz}
\usetikzlibrary{positioning, calc, arrows}

\usepackage{graphics} 
\usepackage{epsfig} 
\usepackage{times} 
\usepackage{amsmath} 
\usepackage{amssymb}  
\usepackage{algorithm}
\PassOptionsToPackage{noend}{algpseudocode}
\usepackage{algpseudocode}
\usepackage[font=small]{caption}
\usepackage{units}

\usepackage{fontenc}
\usepackage{bm}
\usepackage[utf8]{inputenc}
\usepackage{mathtools}
\usepackage{color}

\usepackage{epstopdf}
\usepackage{tabularx}
\usepackage{hhline}
\usepackage{url}
\usepackage{leftidx}
\usepackage{tensor}

\usepackage{gensymb}
\usepackage{booktabs}
\usepackage{flushend}
\usepackage{gensymb}
\usepackage[nolist,nohyperlinks]{acronym}

\begin{document}
\begin{frontmatter}

\title{Linear vs Nonlinear MPC for Trajectory Tracking Applied to Rotary Wing Micro Aerial Vehicles} 

\thanks[footnoteinfo]{This project has received funding from the European Union’s Horizon 2020 research and innovation programme under grant agreement No 644128 and from the Swiss State Secretariat for Education, Research and Innovation (SERI) under contract number 15.0044.}

\author[First]{Mina Kamel, Michael Burri and Roland Siegwart}

\address[First]{Authors are with the Autonomous Systems Lab, ETH Z\"{u}rich, Switzerland, {\tt\small fmina@ethz.ch}}

\begin{abstract}                
Precise trajectory tracking is a crucial property for \acp{MAV} to operate in cluttered environment or under disturbances.
In this paper we present a detailed comparison between two state-of-the-art model-based control techniques for \ac{MAV} trajectory tracking.
A classical \ac{LMPC} is presented and compared against a more advanced \ac{NMPC} that considers the full system model.
In a careful analysis we show the advantages and disadvantages of the two implementations in terms of speed and tracking performance.
This is achieved by evaluating hovering performance, step response, and aggressive trajectory tracking under nominal conditions and under external wind disturbances.
\end{abstract}

\begin{keyword}
UAVs, Predictive Control, Trajectory Tracking and Path Following, Real-time control. 
\end{keyword}

\end{frontmatter}

\section{INTRODUCTION}
\acp{MAV} are gaining a growing attention recently thanks to their agility and ability to perform tasks that humans are unable to do.
Many researches have employed \acp{MAV} successfully to perform infrastructure inspection as shown in \cite{bircher2016receding}, exploration tasks in unknown environment as in \cite{nbvp2016}, search and rescue operations as presented in \cite{oettershagen2016long} or forest resources monitoring as shown in \cite{treecavity2016}.
Precise trajectory tracking is an important feature for aerial robots when operating in real environment under external disturbances that may heavily affect the flight performance, especially in vicinity to structures.
Furthermore, there is a big boost in personal drones which need to be able to track a moving object and take nice aerial footage requiring fast and agile trajectory tracking.

Currently, it is possible to perform mapping, 3D reconstruction, localization, planning and control completely on-board thanks to the great advances in electronics and semiconductor technology. To fully exploit the robot capabilities and to take advantage of the available computation power, optimization-based control techniques are becoming suitable for real-time \ac{MAV} control. 

In this paper, we present a comparison between two of the state-of-the-art trajectory tracking controllers. A \ac{LMPC} based on a linearized model of the \ac{MAV} and a more advanced \ac{NMPC} that considers the full system dynamics. The goal of this comparison is to emphasis the benefits and drawbacks of considering the full system dynamics in terms of performance improvement, disturbance rejection and computation effort.

The general control structure followed in this paper is a cascade control approach, where a reliable and system-specific low-level attitude controller is present as inner-loop, while a model-based trajectory tracking controller is running as an outer-loop. This cascade approach is justified by the fact that critical flight control algorithm is running on a separate navigation hardware, typically based on miro-controller such as PixHawk by \cite{pixhawk}, while high level tasks are typically executed on a more powerful on-board computer. This introduces a separation layer to keep critical code running despite any failure in the high-level computer. 

To achieve high performance with the cascade approach, it is necessary to account for the inner-loop dynamics in the trajectory tracking controller. Therefore, a system identification has been performed on the MAV to identify the closed loop attitude dynamics. 
%
\begin{figure}[t]
	\centering
	\includegraphics[width=0.99\columnwidth]{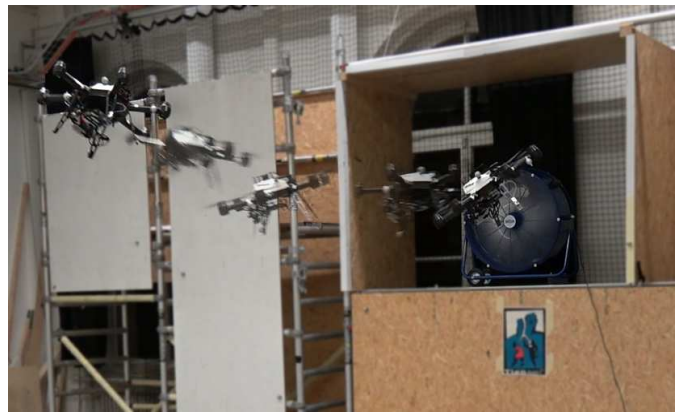}
	\caption{Sequence of the \ac{MAV} poses during aggressive trajectory tracking experiment under wind disturbances.}
	\label{fig:IntroPhoto}
\end{figure}
%

\subsection*{Paper Contributions}
The main goal of this paper is to perform a thorough comparison between a \ac{LMPC} and \ac{NMPC} controller enabling dynamic trajectory tracking for \acp{MAV} together with an open source C++ implementation of both controllers on \cite{mav_control_github}. 
The comparison aims to highlight the benefits of full system dynamics consideration and its effect on the flight envelope.

The remainder of this paper is organized as follows: In Section \ref{sec:related_work} the related work of the problem of MAV trajectory tracking control is discussed. In Section \ref{sec:mav_model} the MAV model is presented. The Linear and Nonlinear Model Predictive Controller are presented in Section \ref{sec:linear_mpc} and Section \ref{sec:nonlinear_mpc} respectively. The external disturbance observer is briefly discussed in Section~\ref{sec:external_disturbances_estimation}. Finally, experimental results are presented and discussed in Section \ref{sec:results}. 

\section{Related Work}\label{sec:related_work}
Many researchers have been focusing on the control problem for MAVs. Among the first controller evaluation on \ac{MAV} is the work done by Samir Bouabdallah in \cite{bouabdallah2004pid} where a comparison between classic PID controller and LQ controller has been conducted for attitude stability.  Surprisingly the authors found that a simple PID controller was performing better than an LQ controller for attitude stability. The authors justify this result as a result of imperfect model.

A commonly used nonlinear controller was proposed by \cite{lee2010geometric}, where the attitude error is directly calculated on the manifold to have a globally stable controller.
Using a very similar formulation, impressive trajectory tracking results were obtained using high gain feedback control \cite{mellinger2012trajectory}.
These controllers, however, are not able to guarantee any state or input constraints and the trajectory needs to be selected carefully.
\ac{MPC} on the other hand is able to directly include constraints in the optimization, which is important for real systems with physical constraints.

In our previous work presented in \cite{kamel2015fast} a \ac{NMPC} approach is successfully employed to control the \ac{MAV} attitude on the Special Orthogonal group $SO(3)$. The controller is based on a geometric formulation of the attitude error. Experimental results showed the ability of the controller to recover from any configuration and  to handle propeller failure. However, the computation cost limits the use of such controller in practice.

In \cite{Bemporad200914} the authors employed a hierarchical \ac{MPC} to achieve stability and autonomous navigation for MAVs. To achieve stable flight, a linear \ac{MPC} is employed while a hybrid \ac{MPC} approach is used to generate collision-free trajectory which is tracked by the linear \ac{MPC}. However, the proposed stabilizing controller has not been tested experimentally and controller performance is not evaluated under disturbances.

To add robustness with respect to time delays in on-board estimation \cite{blosch2010vision} proposed to separate the controller into an attitude controller running on the micro controller and an \ac{LQR} for position tracking.
For better tracking performance the closed loop dynamics of the attitude controller was identified and included in the \ac{LQR}
This approach was extended in \cite{burri2012aerial} with a \ac{LMPC} for position tracking.
In this work we extended this controller with a simple aerodynamic drag model and use it as our reference implementation.

A similar formulation was also used in \cite{Alexis2016}.
The authors present a Robust \ac{MPC} approach to stabilize the vehicle and demonstrate the effectiveness of the proposed controller under wind disturbances and slung load. 

The authors in \cite{berkenkamp-ecc15} propose a novel approach to deal with conservativeness of robust control techniques. The system uncertainty is learned online using Gaussian Process and the linear robust controller is updated according to the learned uncertainties achieving less conservative linear robust controller.

\section{MAV Model}\label{sec:mav_model}
In this section we present the MAV model employed in the controllers formulation. We first introduce the full vehicle model and explain the forces and moment acting on the system. Next, we will briefly discuss the closed-loop attitude model employed in the trajectory tracking controller.
 
\subsubsection{System model}
\begin{figure}[t]
 	\centering
	\resizebox{0.99\columnwidth}{!}{%
		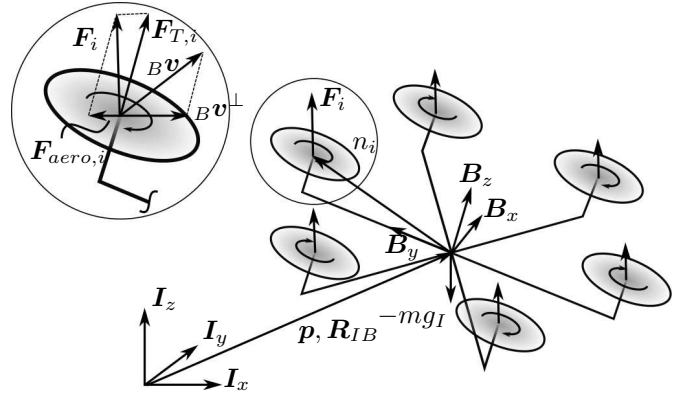
	}%
	\caption{A schematic of \ac{MAV} showing Forces and torques acting on the \ac{MAV} and aerodynamic forces acting on a single rotor. Inertial and \ac{CoG} frames are also shown.}
 	\label{fig:mav_frames}
 \end{figure}

We define the world fixed inertial frame $ I $ and the body fixed frame $ B $ attached to the \ac{MAV} in the \ac{CoG} as shown in Figure~\ref{fig:mav_frames}. The vehicle configuration is described by the position of the \ac{CoG} in the inertial frame $ \bm{p} \in \mathbb{R}^{3} $, the vehicle velocity in inertial frame $ \bm{v} $, the vehicle orientation  $ \bm{R}_{I B}\in SO(3) $  and the body angular rate $ \bm{\omega} $.

The main forces acting on the vehicle are generated from the propellers. Each propeller generates thrust proportional to the square of the propeller rotation speed and angular moment due to the drag force. The generated thrust $ \bm{F}_{T,i} $ and moment $ \bm{M}_{i} $ from the $ i-th $ propeller is given by:

\begin{subequations}
	\begin{align}
	\bm{F}_{T,i} &= k_{n} n_{i}^{2} \bm{e}_z, 				\label{eq:prop_force}\\
	\bm{M}_{i} &= (-1)^{i-1} k_{m} \bm{F}_{T,i}, 		\label{eq:prop_moment}
	\end{align}
\end{subequations}
where $ n_{i} $ is the $ i-th $ rotor speed, $ k_n $ and $ k_m $ are positive constants and $\bm{e}_z$ is a unit vector in $z$ direction.

Moreover, we consider two important effects that appear in case of dynamic maneuvers. These effects are the blade flapping and induced drag. The importance of these effects stems from the fact that they introduce additional force in the $ x-y $ rotor plane adding more damping to the \ac{MAV} as shown in \cite{6289431}. It is possible to combine these effects as shown in \cite{6696696} into one lumped drag coefficient $k_D$.

This leads to the aerodynamic force $\bm{F}_{aero,i}$:
 \begin{equation}
 \bm{F}_{aero,i} =  f_{T,i}\bm{K}_{drag}\bm{R}_{IB}^{T}\bm{v}
 \end{equation}
 where $  \bm{K}_{drag} = diag(k_D,k_D,0)$ and $f_{T,i}$ is the $z$ component of the $ i-th $ thrust force.
 
 \graphicspath{{Figures/}}
 %

 %
 
The motion of the vehicle can be described by the following equations:
\normalsize
\begin{subequations}
\begin{eqnarray}
 \dot{\bm{p}} &=& \bm{v} ,\label{eq:dynamics_eq1} \\
 \dot{\bm{v}} &=& \frac{1}{m}\left( \bm{R}_{I B} \sum_{i=0}^{N_r}\bm{F}_{T,i} - \bm{R}_{IB}\sum_{i=0}^{N_r} \bm{F}_{aero,i}  + \bm{F}_{ext}\right) \nonumber \\ 
 & +& \left[ \begin{array}{ccc}
 0 \\ 
 0 \\ 
 -g
 \end{array}  \right],  \label{eq:dynamics_eq2} \\
 \dot{\bm{R}}_{IB} &=& \bm{R}_{IB} \lfloor \bm{\omega} \times \rfloor,  \label{eq:dynamics_eq3}\\
 \bm{J} \dot{\bm{\omega}}& = & -\bm{\omega} \times \bm{J} \bm{\omega} + \bm{\mathcal{A}} \left[ \begin{array}{c}
 n^2_{1} \\ 
 \vdots \\ 
 n^2_{N_{r}}
 \end{array}  \right], \label{eq:dynamics_eq4}
 \end{eqnarray}
\end{subequations}
\normalsize
\noindent where $ m $ is the mass of the vehicle and $ \bm{F}_{ext} $ is the external forces acting on the vehicle (i.e wind). $ \bm{J} $ is the inertia matrix, $ \mathbf{\mathcal{A}}  $ is the allocation matrix and $N_{r}$ is the number of propellers.

\subsubsection{Attitude model}
We follow a cascaded approach as described in \cite{blosch2010vision} and assume that the vehicle attitude is controlled by an attitude controller.
For completeness we quickly summarize the findings in the following.

To achieve accurate trajectory tracking, it is crucial for the high level controller to consider the inner loop system dynamics.
Therefore, it is necessary to consider a simple model of the attitude closed-loop response.
These dynamics can either be calculated by simplifying the closed loop dynamic equations (if the controller is known) or by a simple system identification procedure in case of an unknown attitude controller (on commercial platforms for instance).
In this work we used the system identification approach to identify a first order closed-loop attitude response. 

The inner-loop attitude dynamics are then expressed as follows:

\begin{subequations}
	\begin{align}
	&\dot{\phi} = \frac{1}{\tau}_{\phi} \left(k_{\phi} \phi_{cmd} - \phi  \right), \label{eq:roll_dynamics} \\ 
	&\dot{\theta} = \frac{1}{\tau}_{\theta} \left( k_{\theta}\theta_{cmd} - \theta \right),\label{eq:pitch_dynamics}\\ 
	&\dot{\psi} = \dot{\psi}_{cmd}  \label{eq:yaw_dynamics}
	\end{align}
	\label{eq:attitude_innerloop}
\end{subequations}
where $ k_{\phi}, k_{\theta} $ and $ \tau_{\phi}, \tau_{\theta} $ are the gains and time constant of roll and pitch angles respectively. $ \phi_{cmd} $ and $ \theta_{cmd} $ are the commanded roll and pitch angles and $ \dot{\psi}_{cmd} $ is commanded angular velocity of the vehicle heading.

The aforementioned model will be employed in the subsequent trajectory tracking controllers to account for the inner-loop dynamics. Note that the vehicle heading angular rate $ \dot{\psi} $  is assumed to track the command instantaneously. This assumption is reasonable because the \ac{MAV} heading angle has no effect on the \ac{MAV} position.

\section{Linear MPC}\label{sec:linear_mpc}
In this section, we describe a Model Predictive Controller to achieve trajectory tracking using a simple model of the MAV \cite{mpcbook}. In Subsection~\ref{subsec:model_linearization}, the vehicle model is linearized around hovering condition. Finally, in Subsection~\ref{subsec:lmpc_formulation}, we formulate the \ac{LMPC} controller and discuss nonlinearity compensation. 
\subsection{Model Linearization}\label{subsec:model_linearization}
The vehicle model can be linearize around hovering condition assuming small attitude angles and vehicle heading aligned with the first axis of inertial frame ($ \psi = 0 $). 
We define the following state vector:
\begin{equation}\label{eq:state_vector}
\bm{x} = \left(\begin{array}{ccccc}
\bm{p}^{T}  & \bm{v}^{T}  & {}_{I}\phi & {}_{I}\theta
\end{array}  \right) ^{T},
\end{equation}
and control input vector:
\begin{equation}\label{eq:control_input_vector}
\bm{u} = \left(\begin{array}{ccc}
{}_{I}\phi_{cmd}  & {}_{I}\theta_{cmd}  & T_{cmd}
\end{array}  \right) ^{T} 
\end{equation}
where $ T_{cmd} $ is the commanded thrust, which we assume can be achieve instantaneously as the motors dynamics are typically very fast. $ {}_{I}\phi,  {}_{I}\theta $ are the roll and pitch angles which we denote in inertial frame to get rid of the vehicle heading $ \psi $ from the model. The transformation between attitude angles and heading free attitude angles is given by:
\begin{equation}
\label{eq:rotation_cmd_z}
\begin{pmatrix}
\phi \\
\theta
\end{pmatrix} = \begin{pmatrix}
\cos{\psi} & \sin{\psi} \\
-\sin{\psi} & \cos{\psi}
\end{pmatrix}\begin{pmatrix}
{}_{I}\phi \\
{}_{I}\theta
\end{pmatrix}.
\end{equation}

After linearization and discretization of the model described in Section~\ref{sec:mav_model}, the following linear state-space model is obtained:

\begin{equation}\label{eq:sys_linear}
\bm{x}_{k+1} =\bm{A} \bm{x}_{k} + \bm{B}\bm{u}_{k} + \bm{B}_{d}\bm{F}_{ext, k}.
\end{equation}


\subsection{Controller Formulation}\label{subsec:lmpc_formulation}
The \ac{LMPC} scheme repeatedly solves the following \ac{OCP}: 
\small
\begin{equation} \label{eq:mav_linear_mpc_opt}
\begin{aligned}
\min_{\bm{U},\bm{X}} 
\sum_{k=0}^{N-1} &\left( \left\|(\bm{x}_{k} - \bm{x}_{ref, k}\right\|_{\bm{Q}_{x}}^{2} 
+  \left\|\bm{u}_{k} - \bm{u}_{ref, k}\right\|_{\bm{R}_{u}}^{2}\right) \\
& +  \left\|\bm{x}_{N} - \bm{x}_{ref, N}\right\|^{2}_{\bm{P}}\\
&\begin{aligned}
\text{subject to} &
& & \bm{x}_{k+1} = \bm{A}\bm{x}_{k} + \bm{B}\bm{u}_{k} + \bm{B}_{d}\bm{F}_{ext,k};\\
& & & \bm{F}_{ext,k+1} = \bm{F}_{ext,k}, \quad k=0, \dots, N-1\\
& & & \bm{u}_{k} \in \mathbb{U} \\
& & & \bm{x}_{0} = \bm{x}\left( {t_0}\right), \quad \bm{F}_{ext,0} = \bm{F}_{ext}\left(t_{0} \right) .
\end{aligned}
\end{aligned}
\end{equation}
\normalsize

\noindent where $ \bm{Q}_{x} \succeq 0 $ is the penalty on the state error, $ \bm{R}_{u} \succ 0 $ is the penalty on control input error and $ \bm{P}  \succeq 0 $ is the terminal state error penalty. $ \bm{x}_{ref, k} $ and $ \bm{u}_{ref, k} $ are respectively the target state vector and target control input at time $ k $.  $ \mathbb{U} $ is the control input constraint given by:
\small
\begin{equation}\label{eq:input_constraints_lmpc}
\mathbb{U} = \left\lbrace \bm{u} \in \mathbb{R}^{3} | \left[ \begin{array}{c}
\phi_{min} \\ 
\theta_{min} \\ 
T_{cmd,min}
\end{array} \right] \leq \bm{u} \leq   \left[ \begin{array}{c}
\phi_{max} \\ 
\theta_{max} \\ 
T_{cmd,max}
\end{array} \right] \right\rbrace 
\end{equation} 
\normalsize

Note that we assume constant disturbances along the prediction horizon. Only the first control input $ \bm{u}_{0} $ is applied to the system, and the process is repeated the next time step in a receding horizon fashion.

The attitude commands $ _{I}\phi_{cmd} , _{I}\theta_{cmd}$ is transformed into the \ac{MAV} body frame by applying \eqref{eq:rotation_cmd_z}.  Moreover, the non-zero \ac{MAV} attitude effects on the generated lift can be compensated. The actual thrust control input to the \ac{MAV} low-level attitude controller is given by:

\begin{equation}
\begin{aligned}
\tilde{T}_{cmd} &= \frac{T_{cmd} + g}{\cos{\phi} \cos{\theta}} \\
\end{aligned}
\label{eq:thrust_nonlinear_compensation}
\end{equation}

To achieve a better tracking performance of dynamic trajectories, we can include a feed-forward term by setting the reference control input $ \bm{u}_{ref} $ as follows: 

\begin{equation}
 \bm{u}_{ref, k} = \left( \begin{array}{ccc}
 -_{B}\ddot{y}_{ref, k} / g & _{B}\ddot{x}_{ref, k} / g & _{B}\ddot{z}_{ref}
 \end{array} \right)^{T}
\end{equation}

\noindent where $  _{B}\ddot{x}_{ref, k}, \;   _{B}\ddot{y}_{ref, k},  \; _{B}\ddot{z}_{ref, k} $ are the desired trajectory acceleration expressed in \ac{MAV} body frame $ B $.

\section{Nonlinear  MPC}\label{sec:nonlinear_mpc}
In this section, a continuous time  \ac{NMPC} controller that considers the full system dynamics explained in Section~\ref{sec:mav_model} is formulated.  In Subsection~\ref{subsec:nmpc_controller_formulation} we formulate the \ac{OCP} and in Subsection~\ref{subsec:ocp_solution} we discuss the technique employed to solve the \ac{OCP} to achieve real-time implementation.

\subsection{Controller Formulation}\label{subsec:nmpc_controller_formulation}
To formulate the \ac{NMPC}, we first define the following state vector:
\begin{equation}\label{eq:state_vector_nmpc}
\bm{x} = \left(\begin{array}{cccccc}
\bm{p}^{T}  & \bm{v}^{T}  & \phi & \theta & \psi
\end{array}  \right) ^{T},
\end{equation}
and control input vector:
\begin{equation}\label{eq:control_input_vector_nmpc}
\bm{u} = \left(\begin{array}{ccc}
\phi_{cmd}  & \theta_{cmd}  & T_{cmd}
\end{array}  \right) ^{T} 
\end{equation}

Now, we can define the \ac{OCP} as follows:

\small
\begin{equation} \label{eq:mav_nonlinear_mpc_opt}
\begin{aligned}
\min_{\bm{U},\bm{X}} &\
\int_{t=0}^{T} \left\|\bm{x}(t) - \bm{x}_{ref}(t)\right\|^{2}_{\bm{Q}_{x}} + \left\|\bm{u}(t) - \bm{u}_{ref}(t)\right\|^{2}_{\bm{R}_{u}}dt \\ & + \left\|\bm{x}(T) - \bm{x}_{ref}(T)\right)\|^{2}_{\bm{P}}\\
&\begin{aligned}
\text{subject to} &
& & \dot{\bm{x}} = \bm{f}(\bm{x}, \bm{u});\\
& & & \bm{u}(t) \in \mathbb{U} \\
& & & \bm{x}(0) = \bm{x}\left( {t_0}\right).
\end{aligned}
\end{aligned}
\end{equation}
\normalsize
\noindent where $ \bm{f} $ is composed by Equations~\eqref{eq:dynamics_eq1},~\eqref{eq:dynamics_eq2} and~\eqref{eq:attitude_innerloop}.
The controller is implemented in a receding horizon fashion, where the aforementioned optimization problem needs to be solved every time step in and only the first control input is actually applied to the system. Solving \eqref{eq:mav_nonlinear_mpc_opt} repeatedly in real-time is not a trivial task.  \emph{Direct methods}  techniques has gained a particular attention recently to address optimal control problems. In the next subsection we describe the method employed by the solver to solve \eqref{eq:mav_nonlinear_mpc_opt}. 

\subsection{\ac{OCP} Solution}\label{subsec:ocp_solution}

\emph{Multiple shooting} technique is employed to solve \eqref{eq:mav_nonlinear_mpc_opt}. The system dynamics and constraints are discretized over a coarse discrete time grid $ t_{0} , \dots, t_{N}$ within the time interval $ \left[ t_{k}, t_{k+1} \right]  $. For each interval, a \ac{BVP} is solved, where  additional continuity constrains are imposed. An implicit \emph{RK} integrator of order $ 4 $ is employed to forward simulate the system dynamics along the interval.
At this point, the \ac{OCP} can be expressed as a \ac{NLP} that can be solved using \ac{SQP} technique where an active set or interior point method can be used to solve the \ac{QP}.

\section{External Disturbances Estimation}\label{sec:external_disturbances_estimation}
In this section, we discuss the external disturbances estimator employed to achieve offset-free trajectory tracking.
The external disturbances $ \bm{F}_{ext} $ is estimated by an augmented state \ac{EKF}  that includes external forces. The \ac{EKF} is employing the same model used in the controller design, but it includes also vehicle heading angle $ \psi $. In this way, the external forces will capture also modeling error achieving zero steady state tracking error \cite{mpcbook}. The filter is employing the inner-loop attitude dynamics presented in Equation \eqref{eq:attitude_innerloop}.

\section{Implementation and Results}\label{sec:results}
In this section, we present the results of multiple experiments to evaluate the performance of the previously presented \ac{LMPC} and \ac{NMPC}.

\subsection{System Description}
The aforementioned \ac{LMPC} and \ac{NMPC} have been implemented and evaluated on Asctec NEO hexacopter equipped with an  \emph{Intel 3.1 GHz i7 Core} processor, $ 8 $ GB RAM. The \ac{MAV} parameters are shown in Table \ref{tab:neo_params}. The on-board computer is running \ac{ROS}. The vehicle state is estimated using an \ac{EKF} as described in \cite{lynen2013robust} by fusing vehicle \ac{IMU} with external motion capture system (Vicon).

\subsection{Controllers Implementation}
To implement the \ac{LMPC}, an efficient \ac{QP} solver using CVXGEN code generator framework by \cite{mattingley2012cvxgen}  is employed to solve the optimization problem \eqref{eq:mav_linear_mpc_opt} every time step. While the \ac{NMPC} is implemented by solving \eqref{eq:mav_nonlinear_mpc_opt} every time step using an efficient solver generated using \emph{ACADO} toolkit by \cite{Houska2011a}. A real time iteration scheme based on Gauss-Newton is employed to approximate the optimization problem and iteratively improve the solution during the runtime of the process. In this way, an improvement of the computation time is achieved.
Both controllers are running at $ 100 $ Hz while internally the prediction is performed at $ 10 Hz $, in this way we achieve longer prediction horizon with less computational efforts. The prediction horizon is chosen to be $ 20 $ steps resulting into $ 2 $ seconds prediction horizon.  The penalties for both controllers are chosen such that the cost functions are identical, in this way we emphasis the benefits of full dynamics model over a linearized model. The terminal cost in both controllers is chosen to account for the infinite horizon cost.

\subsection{Hovering Performance}
\begin{figure}[t]
\centering
\includegraphics[width=0.95\linewidth]{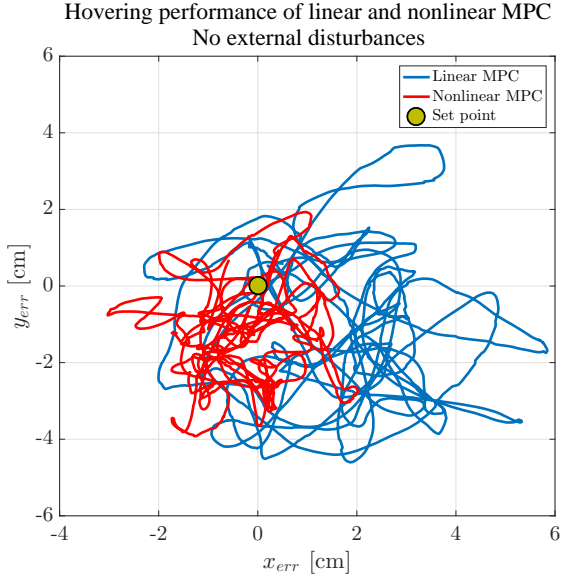}
\caption{$ x-y $ error during hovering using \ac{LMPC} and \ac{NMPC}. No external wind is applied.}
\label{fig:hoveringnowind}
\end{figure}

\begin{figure}[t]
	\centering
	\includegraphics[width=0.95\linewidth]{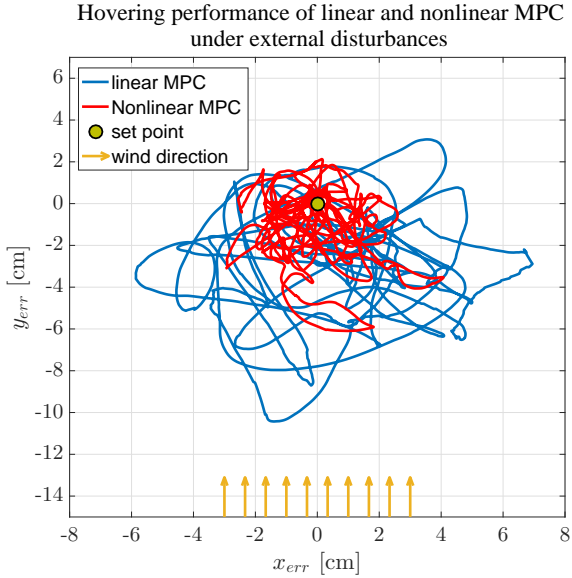}
	\caption{$ x-y $ error during hovering using \ac{LMPC} and \ac{NMPC}. An external wind of around \unit[11]{m/s} is applied in the direction indicated by yellow arrows.}
	\label{fig:hoveringwind}
\end{figure}

\begin{table}[t]
	\centering
	\captionsetup{width=\columnwidth}
	\caption{\ac{RMSE} of individual axes of \ac{LMPC} and \ac{NMPC} during hovering experiment with and without external disturbances. Error is in \unit{cm}.}
	\begin{tabular}{lcc}  \toprule 
		& \ac{LMPC} & \ac{NMPC} \\  \midrule
		without wind &  $\left( \begin{array}{ccc}
		0.9 & 1.4 & 0.7
		\end{array}\right)$ &  $\left( \begin{array}{ccc}
		1.0 & 1.7 & 0.5
		\end{array}\right)$\\ 
		with wind & $\left( \begin{array}{ccc}
		1.6 & 1.9 & 1.1
		\end{array}\right)$   & $\left( \begin{array}{ccc}
		1.2 & 2.0 & 0.9
		\end{array}\right)$   \\ \bottomrule
	\end{tabular} 
	\label{tab:hovering_performance}
\end{table}

\begin{table}[t]
	\centering
	\captionsetup{width=\columnwidth}
	\caption{NEO hexacopter parameters and control input constraints.}
	\begin{tabular}{l|r}  \toprule
		  Parameter & Value \\  \midrule
		 mass & \unit[$3.42$]{kg} \\
		 $\tau_{\phi}$ & \unit[$0.1901$]{sec} \\
		 $\tau_{\theta}$ & \unit[$0.1721$]{sec} \\
		 $k_{\phi}$ & $ 0.91 $ \\
		 $k_{\theta}$ & $ 0.96 $ \\
		 $ \phi_{max}, \theta_{max} $ & \unit[$ 45 $]{\degree} \\ 
		 $ \phi_{min}, \theta_{min} $ & \unit[$ -45 $]{\degree} \\
		 $T_{cmd, max}$ &  \unit[$ 40.3 $]{N} \\
		 $T_{cmd, min}$ & \unit[$ 13.5 $]{N}  \\
		 \bottomrule
	\end{tabular} 
	\label{tab:neo_params}
\end{table}

The purpose of these experiments is to evaluate the hovering performance of the \ac{MAV} in nominal conditions and under external wind disturbances. Figure~\ref{fig:hoveringnowind} shows the $ x, y $ error under no external disturbances for \ac{LMPC} and \ac{NMPC}. Clearly the performance of both controllers is very comparable, and this result is expected since the \ac{LMPC} is employing a model linearized around hovering condition. The \ac{RMSE} is found to be \unit[1.84]{cm} and \unit[2.05]{cm} in the case of \ac{LMPC} and \ac{NMPC} respectively. 

Under external wind of around  \unit[11]{m/s}, the $ x-y $ error is shown in Figure~\ref{fig:hoveringwind}. Both controllers achieve an \ac{RMSE} of \unit[2.7]{cm} and \unit[2.5]{cm} in the case of \ac{LMPC} and \ac{NMPC} respectively. The \ac{RMSE} is shown in Table~\ref{tab:hovering_performance}.

\subsection{Step Response}

\begin{figure}[t]
	\centering
	\includegraphics[width=0.99\linewidth]{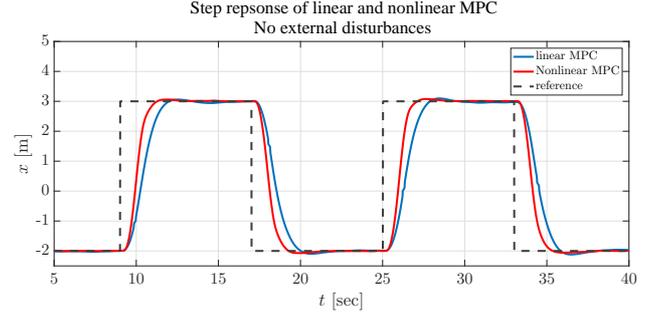}
	\caption{Step response in $ x $ direction.}
	\label{fig:steps}
\end{figure}

\begin{figure}[t]
	\centering
	\includegraphics[width=0.99\linewidth]{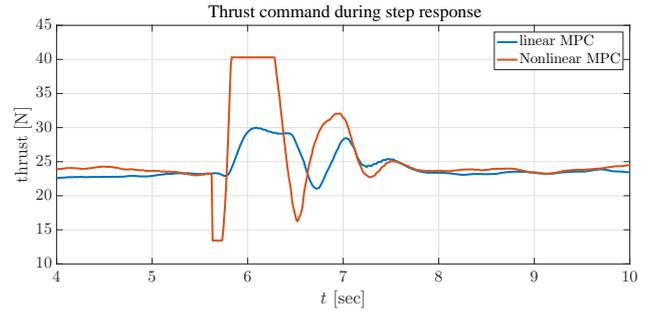}
	\caption{Thrust command during step response.}
	\label{fig:thrust_cmd_step}
\end{figure}

Even though the tracking error is weighted in the same way in both controllers, the \ac{NMPC} outperforms the \ac{LMPC} in the step response as can be seen in Figure~\ref{fig:steps}. The main reason for the is the exploitation of the full system dynamics, in particular, the thrust command coupling with lateral motion. This leads to faster response with no noticeable overshoot. Figure~\ref{fig:thrust_cmd_step} shows the thrust command during step response. The \ac{NMPC} exploits the full thrust command between $ T_{cmd, min} $ and $ T_{cmd, max} $ while the \ac{LMPC} is employing the thrust command in a very limited way, which is introduced by the nonlinear compensation shown in Equation~\eqref{eq:thrust_nonlinear_compensation}. 

To perform qualitative comparison, we compare the rise time and overshoot percentage. The rise time is around \unit[1.6]{s} and \unit[1.0]{s} for the \ac{LMPC} and \ac{NMPC} respectively. While the overshoot percentage is found to be  \unit[1.98]{\%} for both controllers. This clearly highlights that the \ac{NMPC} outperforms the \ac{LMPC} in this case. 

\subsection{Aggressive Trajectory Tracking}

\begin{figure}[t]
	\centering
	\includegraphics[width=0.99\linewidth]{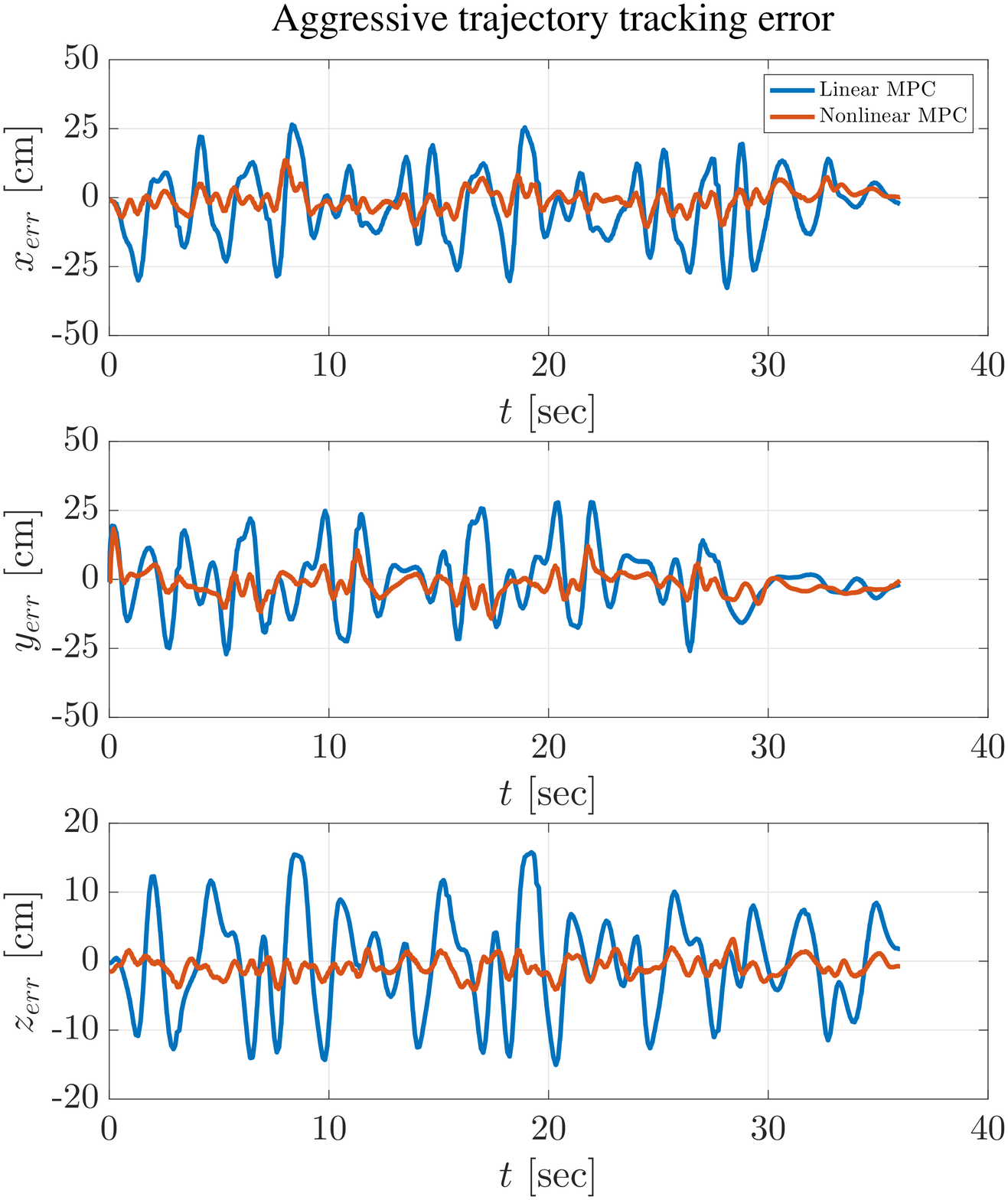}
	\caption{Aggressive trajectory error plots under external wind disturbances. Wind speed is measured to be around \unitfrac[11]{m}{s}.}
	\label{fig:aggressive_traj_err}
\end{figure}

\begin{figure}[t]
	\centering
	\includegraphics[width=0.99\linewidth]{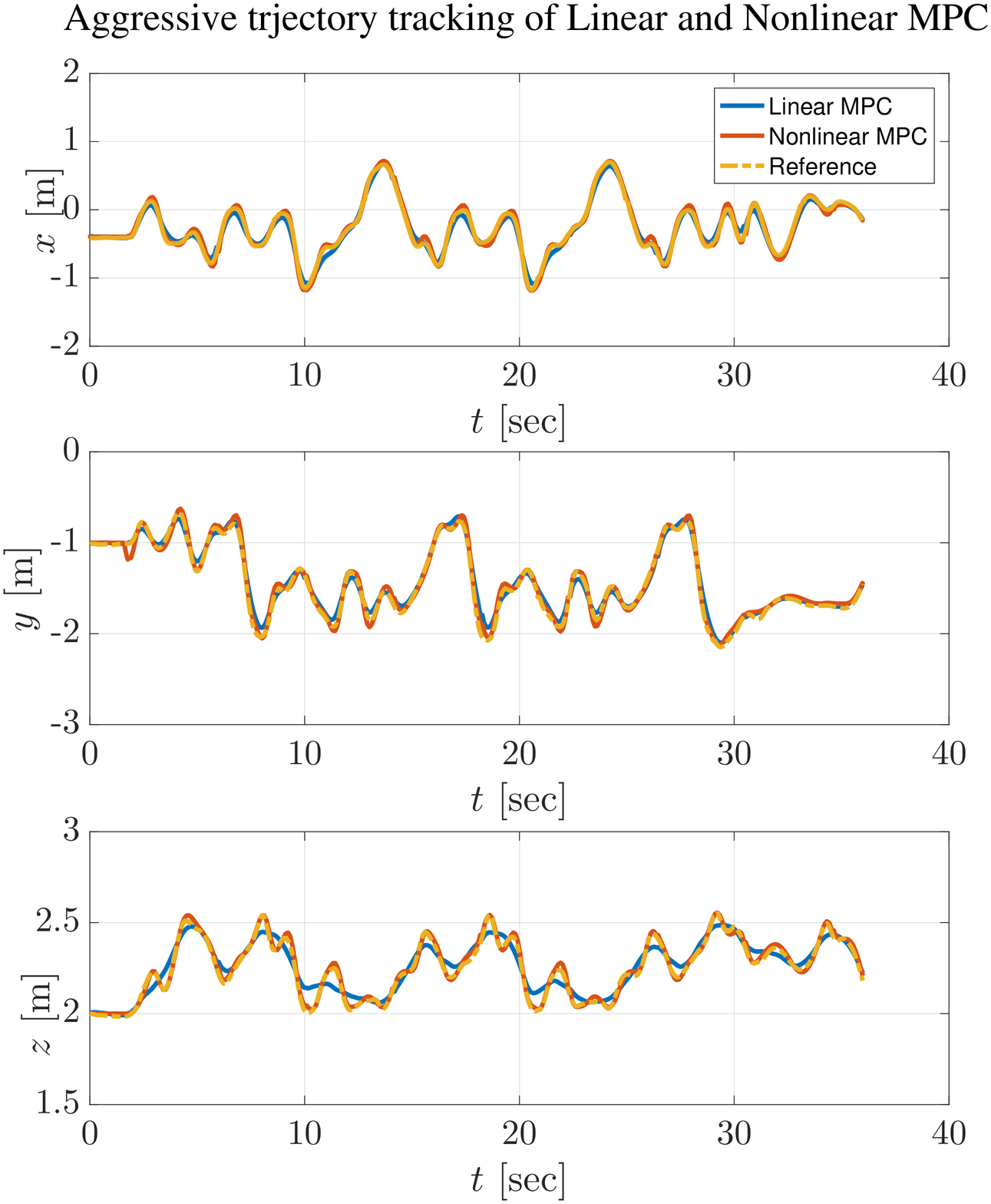}
	\caption{Aggressive trajectory tracking plots under external wind disturbances.}
	\label{fig:aggressive_traj_tracking}
\end{figure}

In this experiment, we compare both controllers capabilities to track an aggressive polynomial trajectory under external disturbances. The trajectory tracking error of both controllers is shown in Figure~\ref{fig:aggressive_traj_err}, while the actual trajectory tracking is shown in Figure~\ref{fig:aggressive_traj_tracking}. The \ac{RMSE} is found to be \unit[10.8]{cm} and \unit[7.1]{cm} in the case of \ac{LMPC} and \ac{NMPC} respectively. 

\begin{figure}[t]
	\centering
	\includegraphics[width=0.99\linewidth]{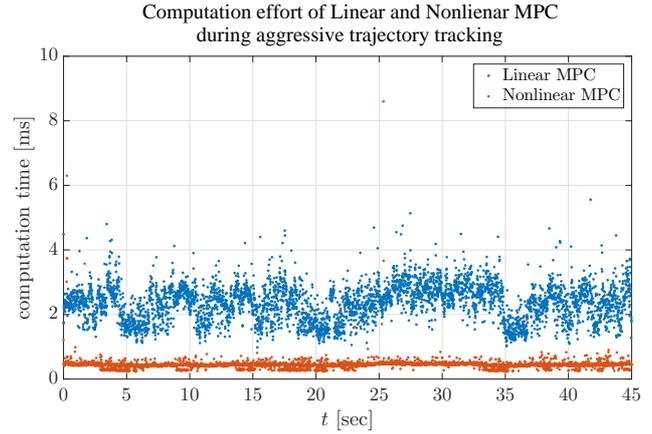}
	\caption{Computation time comparison between \ac{LMPC} and \ac{NMPC} during aggressive trajectory tracking.}
	\label{fig:computation_time}
\end{figure}

To compare the computation effort required by each controller, in Figure~\ref{fig:computation_time} we show the time employed by the solver to find the control action. Clearly, the real time iteration scheme implemented in the \ac{NMPC} greatly improves the solver speed, achieving an average solve time of  \unit[0.45]{ms} for the  \ac{NMPC} compared to \unit[2.35]{ms} for the \ac{LMPC} case, improving the computation effort by a factor of $ 5 $.

\section{CONCLUSION}\label{sec:concl}
In this paper we presented a Linear and Nonlinear Model Predictive Controllers for trajectory tracking of \ac{MAV}. Detailed comparison has been performed during hovering, step response and aggressive trajectory tracking under external disturbances. Both controllers showed comparable behavior while the \ac{NMPC} showed a slightly better disturbance rejection capability, step response, tracking performance and computational effort. An open source implementation of these controllers can be found on \cite{mav_control_github}.

\bibliography{./BIB/mpcbib}

\begin{thebibliography}{21}
\providecommand{\natexlab}[1]{#1}
\providecommand{\url}[1]{\texttt{#1}}
\providecommand{\urlprefix}{URL }
\expandafter\ifx\csname urlstyle\endcsname\relax
  \providecommand{\doi}[1]{doi:\discretionary{}{}{}#1}\else
  \providecommand{\doi}{doi:\discretionary{}{}{}\begingroup
  \urlstyle{rm}\Url}\fi

\bibitem[{Alexis et~al.(2016)Alexis, Papachristos, Siegwart, and
  Tzes}]{Alexis2016}
Alexis, K., Papachristos, C., Siegwart, R., and Tzes, A. (2016).
\newblock Robust model predictive flight control of unmanned rotorcrafts.
\newblock \emph{Journal of Intelligent {\&} Robotic Systems}, 81(3), 443--469.

\bibitem[{Bemporad et~al.(2009)Bemporad, Pascucci, and Rocchi}]{Bemporad200914}
Bemporad, A., Pascucci, C., and Rocchi, C. (2009).
\newblock Hierarchical and hybrid model predictive control of quadcopter air
  vehicles.
\newblock \emph{\{IFAC\} Proceedings Volumes}, 42(17), 14 -- 19.
\newblock \doi{http://dx.doi.org/10.3182/20090916-3-ES-3003.00004}.
\newblock 3rd \{IFAC\} Conference on Analysis and Design of Hybrid Systems.

\bibitem[{Berkenkamp and Schoellig(2015)}]{berkenkamp-ecc15}
Berkenkamp, F. and Schoellig, A.P. (2015).
\newblock Safe and robust learning control with {G}aussian processes.
\newblock In \emph{{Proc. of the European Control Conference (ECC)}},
  2501--2506.

\bibitem[{Bircher et~al.(2016{\natexlab{a}})Bircher, Kamel, Alexis, Oleynikova,
  and Siegwart}]{nbvp2016}
Bircher, A., Kamel, M., Alexis, K., Oleynikova, H., and Siegwart, R.
  (2016{\natexlab{a}}).
\newblock Receding horizon "next-best-view" planner for 3d exploration.
\newblock In \emph{2016 IEEE International Conference on Robotics and
  Automation (ICRA)}, 1462--1468.
\newblock \doi{10.1109/ICRA.2016.7487281}.

\bibitem[{Bircher et~al.(2016{\natexlab{b}})Bircher, Kamel, Alexis, Oleynikova,
  and Siegwart}]{bircher2016receding}
Bircher, A., Kamel, M., Alexis, K., Oleynikova, H., and Siegwart, R.
  (2016{\natexlab{b}}).
\newblock Receding horizon path planning for 3d exploration and surface
  inspection.
\newblock \emph{Autonomous Robots}, 1--16.

\bibitem[{Bl{\"o}sch et~al.(2010)Bl{\"o}sch, Weiss, Scaramuzza, and
  Siegwart}]{blosch2010vision}
Bl{\"o}sch, M., Weiss, S., Scaramuzza, D., and Siegwart, R. (2010).
\newblock Vision based mav navigation in unknown and unstructured environments.
\newblock In \emph{Robotics and automation (ICRA), 2010 IEEE international
  conference on}, 21--28. IEEE.

\bibitem[{Borrelli et~al.(2015)Borrelli, Bemporard, and Morari}]{mpcbook}
Borrelli, F., Bemporard, A., and Morari, M. (2015).
\newblock \emph{Predictive Control for Linear and Hybrid Systems}.

\bibitem[{Bouabdallah et~al.(2004)Bouabdallah, Noth, and
  Siegwart}]{bouabdallah2004pid}
Bouabdallah, S., Noth, A., and Siegwart, R. (2004).
\newblock Pid vs lq control techniques applied to an indoor micro quadrotor.
\newblock In \emph{Intelligent Robots and Systems, 2004.(IROS 2004).
  Proceedings. 2004 IEEE/RSJ International Conference on}, volume~3,
  2451--2456. IEEE.

\bibitem[{Burri et~al.(2012)Burri, Nikolic, H{\"u}rzeler, Caprari, and
  Siegwart}]{burri2012aerial}
Burri, M., Nikolic, J., H{\"u}rzeler, C., Caprari, G., and Siegwart, R. (2012).
\newblock Aerial service robots for visual inspection of thermal power plant
  boiler systems.
\newblock In \emph{Applied Robotics for the Power Industry (CARPI), 2012 2nd
  International Conference on}, 70--75. IEEE.

\bibitem[{Houska et~al.(2011)Houska, Ferreau, and Diehl}]{Houska2011a}
Houska, B., Ferreau, H., and Diehl, M. (2011).
\newblock {ACADO} {T}oolkit -- {A}n {O}pen {S}ource {F}ramework for {A}utomatic
  {C}ontrol and {D}ynamic {O}ptimization.
\newblock \emph{Optimal Control Applications and Methods}, 32(3), 298--312.

\bibitem[{Kamel(2016)}]{mav_control_github}
Kamel, M. (2016).
\newblock Open source mav mpc controllers.
\newblock \urlprefix\url{https://github.com/ethz-asl/mav_control_rw}.

\bibitem[{Kamel et~al.(2015)Kamel, Alexis, Achtelik, and
  Siegwart}]{kamel2015fast}
Kamel, M., Alexis, K., Achtelik, M., and Siegwart, R. (2015).
\newblock Fast nonlinear model predictive control for multicopter attitude
  tracking on so (3).
\newblock In \emph{Control Applications (CCA), 2015 IEEE Conference on},
  1160--1166. IEEE.

\bibitem[{Lee et~al.(2010)Lee, Leoky, and McClamroch}]{lee2010geometric}
Lee, T., Leoky, M., and McClamroch, N.H. (2010).
\newblock Geometric tracking control of a quadrotor uav on se (3).
\newblock In \emph{49th IEEE conference on decision and control (CDC)},
  5420--5425. IEEE.

\bibitem[{Lynen et~al.(2013)Lynen, Achtelik, Weiss, Chli, and
  Siegwart}]{lynen2013robust}
Lynen, S., Achtelik, M.W., Weiss, S., Chli, M., and Siegwart, R. (2013).
\newblock A robust and modular multi-sensor fusion approach applied to mav
  navigation.
\newblock In \emph{Intelligent Robots and Systems (IROS), 2013 IEEE/RSJ
  International Conference on}, 3923--3929. IEEE.

\bibitem[{Mahony et~al.(2012)Mahony, Kumar, and Corke}]{6289431}
Mahony, R., Kumar, V., and Corke, P. (2012).
\newblock Multirotor aerial vehicles: Modeling, estimation, and control of
  quadrotor.
\newblock \emph{IEEE Robotics Automation Magazine}, 19(3), 20--32.
\newblock \doi{10.1109/MRA.2012.2206474}.

\bibitem[{Mattingley and Boyd(2012)}]{mattingley2012cvxgen}
Mattingley, J. and Boyd, S. (2012).
\newblock Cvxgen: A code generator for embedded convex optimization.
\newblock \emph{Optimization and Engineering}, 13(1), 1--27.

\bibitem[{Meier(2016)}]{pixhawk}
Meier, L. (2016).
\newblock Pixhawk autopilot.
\newblock \urlprefix\url{https://pixhawk.org}.
\newblock Accessed: 2016-11-08.

\bibitem[{Mellinger et~al.(2012)Mellinger, Michael, and
  Kumar}]{mellinger2012trajectory}
Mellinger, D., Michael, N., and Kumar, V. (2012).
\newblock Trajectory generation and control for precise aggressive maneuvers
  with quadrotors.
\newblock \emph{The International Journal of Robotics Research},
  0278364911434236.

\bibitem[{Oettershagen et~al.(2016)Oettershagen, Stastny, Mantel, Melzer,
  Rudin, Gohl, Agamennoni, Alexis, and Siegwart}]{oettershagen2016long}
Oettershagen, P., Stastny, T., Mantel, T., Melzer, A., Rudin, K., Gohl, P.,
  Agamennoni, G., Alexis, K., and Siegwart, R. (2016).
\newblock Long-endurance sensing and mapping using a hand-launchable
  solar-powered uav.
\newblock In \emph{Field and Service Robotics}, 441--454. Springer.

\bibitem[{Omari et~al.(2013)Omari, Hua, Ducard, and Hamel}]{6696696}
Omari, S., Hua, M.D., Ducard, G., and Hamel, T. (2013).
\newblock Nonlinear control of vtol uavs incorporating flapping dynamics.
\newblock In \emph{2013 IEEE/RSJ International Conference on Intelligent Robots
  and Systems}, 2419--2425.
\newblock \doi{10.1109/IROS.2013.6696696}.

\bibitem[{Steich et~al.(2016)Steich, Kamel, Beardsleys, Obrist, Siegwart, and
  Lachat}]{treecavity2016}
Steich, K., Kamel, M., Beardsleys, P., Obrist, M.K., Siegwart, R., and Lachat,
  T. (2016).
\newblock Tree cavity inspection using aerial robots.
\newblock In \emph{2016 IEEE/RSJ International Conference on Intelligent Robots
  and Systems (IROS)}.

\end{thebibliography}

\begin{acronym}
	\acro{MAV}{Micro Air Vehicle}
	\acro{CoG}{Center of Gravity}
	\acro{MPC}{Model Predictive Controller}
	\acro{LMPC}{Linear Model Predictive Controller} 
	\acro{NMPC}{Nonlinear Model Predictive Controller}
	\acro{DOF}{degrees of freedom}
	\acro{EKF}{Extended Kalman Filter}
	\acro{KF}{Kalman Filter}
	\acro{OCP}{Optimal Control Problem}
	\acro{BVP}{Boundary Value Problem}
	\acro{NLP}{Nonlinear Program}
	\acro{SQP}{Sequential Quadratic Programming}
	\acro{QP} {Quadratic Program}
	\acro{LQR}{Linear-Quadratic Regulator}
	\acro{CoG}{Center of Gravity}
	\acro{ROS}{Robot Operating System}
	\acro{IMU}{Inertial Measurement Unit}
	\acro{RMSE}{Root Mean Squared Error}
		
\end{acronym}

\end{document}